%% file: camera-ready.tex
\pdfoutput=1

\documentclass[11pt]{article}

\usepackage[]{acl}
\usepackage{times}
\usepackage{latexsym}

\usepackage[T1]{fontenc}

\usepackage[utf8]{inputenc}

\usepackage{subfigure}
\usepackage{graphics}

\usepackage{microtype}

\usepackage{booktabs} 
\usepackage{latexsym}
\usepackage{amsmath}
\usepackage{xspace}
\usepackage{amssymb}
\usepackage{graphicx}
\usepackage{multirow}
\usepackage{enumitem}
\usepackage{ulem}
\usepackage{float}
\usepackage{makecell}

\DeclareMathOperator{\softmax}{softmax}
\DeclareMathOperator{\gelu}{GELU}
\DeclareMathOperator{\sigmoid}{sigmoid}
\DeclareMathOperator{\argmax}{argmax}

\newcommand{\ie}{\textit{i.e.,}\xspace}
\newcommand{\eg}{\textit{e.g.,}\xspace}

\newcommand{\paratitle}[1]{\vspace{0.8ex}\noindent \textbf{#1}}
\newcommand{\polnear}{PolNeAR\xspace}
\newcommand{\riqua}{Riqua\xspace}
\newcommand{\datasetzh}{PoliticsZH\xspace}
\newcommand{\modelname}{CofeNet\xspace}

\title{CofeNet: Context and Former-Label Enhanced Net for Complicated Quotation Extraction}

\author{Yequan Wang$^{1*}$, Xiang Li$^{2*}$, Aixin Sun$^{3}$, Xuying Meng$^{4}$, Huaming Liao$^{4}$, Jiafeng Guo$^{4}$\\
$^{1}$Beijing Academy of Artificial Intelligence, Beijing, China\\
$^{2}$Alibaba Group, China\\
$^{3}$School of Computer Science and Engineering, Nanyang Technological University, Singapore\\
$^{4}$Institute of Computing Technology, Chinese Academy of Sciences, Beijing, China\\
tshwangyequan@gmail.com, yuanye.lx@alibaba-inc.com, axsun@ntu.edu.sg, \\
\{mengxuying, lhm, guojiafeng\}@ict.ac.cn\\ 
}

\date{}

\begin{document}
\maketitle

\renewcommand{\thefootnote}{\fnsymbol{footnote}}
\footnotetext[1]{Indicates equal contribution}
\renewcommand{\thefootnote}{\arabic{footnote}}

\begin{abstract}
Quotation extraction aims to extract quotations from written text. There are three components in a quotation: \textit{source} refers to the holder of the quotation,  \textit{cue} is the trigger word(s), and \textit{content} is the main body.
Existing solutions for quotation extraction mainly utilize rule-based approaches and sequence labeling models. While rule-based approaches often lead to low recalls, sequence labeling models cannot well handle quotations with complicated structures.
In this paper, we propose the \textbf{Co}ntext and \textbf{F}ormer-Label \textbf{E}nhanced \textbf{Net}~(\modelname) for quotation extraction. \modelname is able to extract complicated quotations with components of variable lengths and complicated structures. On two public datasets (\ie \polnear and \riqua) and one proprietary dataset (\ie \datasetzh), we show that our \modelname achieves state-of-the-art performance on complicated quotation extraction.

\end{abstract}

\section{Introduction}
\label{sec:intro}

\begin{figure}
    \centering
    \hspace{-4mm}\includegraphics[scale=0.5]{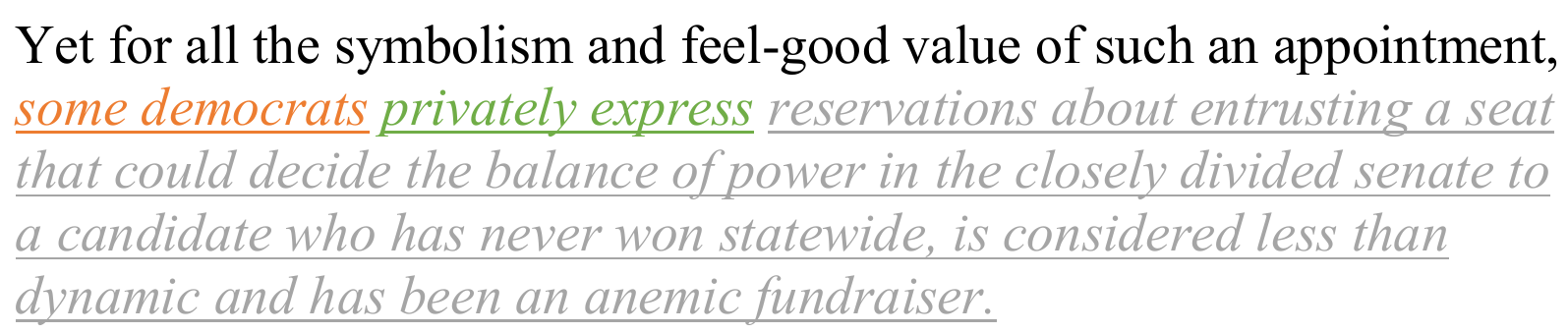}
    \caption{An example of quotations. Text spans with orange, green and gray denote \textit{source}, \textit{cue} and \textit{content} respectively.
    }
    \label{fig:example}
    \vspace{-1.5ex}
\end{figure}

Quotation extraction aims to extract quotations from written text~\cite{pouliquen2007automatic}. For example, given one instance shown in Figure~\ref{fig:example}, we extract the quotation with \textit{source}: \uline{\textit{some democrats}}, \textit{cue}: \uline{\textit{privately express}}, and \textit{content}: \uline{\textit{reservations about ...}}.
As a point of view, quotations provide opinions of the speaker, which is important for analyzing the speaker's stand. In general, quotation extraction is the first step before any further analysis, \eg speaker stand detection. In this paper, we focus on the extraction of the three quotation components.

As illustrated in the above example, the extraction of \textit{content} component in a quotation is complicated and difficult due to three reasons:  variable length, unclear boundary, and indistinguishable components. Specifically, the length of \textit{content} can be over $10$, or even more than $50$ tokens. 
Moreover, \textit{content} does not come with a regular pattern, which not only leads to a more unclear boundary of itself, but also affects the estimation of \textit{source} and \textit{cue}. For example, \textit{content} in a quotation can be a complete instance with subject, predicate, and object. It is therefore hard to distinguish a noun (subject or object) representing the \textit{source} or a part of \textit{content}. Difficulty also exists in recognition of \textit{cue} when tackling with a predicate, \eg verb. 
Thus, as \textit{content} may contain another quotation, such a nesting structure further increases the difficulty of extracting quotations. 

Many existing solutions for quotation extraction are rule-based methods~\cite{pouliquen2007automatic,krestel-etal-2008-minding,Elson2010AutomaticAO,vu-etal-2018-vncorenlp}.
Generally, quotations include direct quotations and indirect quotations. 
Quotation marks and their variants are clear; thus \textit{content} can be extracted by using regular expressions. However, not all quoted texts are quotations. Meanwhile, not all quotations are quoted. Another popular rule-based approach is to recognize \textit{cue} words, \eg \textit{speak(s)}. Similarly, not all \textit{cue} words are related to quotations and vice versa. For both approaches, after recognizing \textit{content} or \textit{cue}, they usually search for the nearby noun as \textit{source}. In short, rule-based methods only cover limited cases, leading to serious low recall problems.

Quotation extraction has also been formulated as a sequence labeling task.
\citet{pareti-etal-2013-automatically,DBLP:conf/www/LeeCCH20} directly adopt sequence labeling for quotation extraction. However, these solutions ignore the traits of quotations where lengths of quotation components are variable and  structures of \textit{content} are complicated. In general, \textit{source} and \textit{cue} components are short, \eg $\leq 3$ tokens.  However, \textit{content} usually is over $10$ tokens, or even more. Further, the complicated structure of \textit{content} greatly reduces the performance of \textit{content} extraction for sequence-labeling-based solutions.

In this paper, we propose \textbf{Co}ntext and \textbf{F}ormer-Label \textbf{E}nhanced \textbf{Net}~(\modelname) for quotation extraction. \modelname is a novel architecture to extract quotations with variable-length and complicated-structured components. Our model is also capable of extracting both direct and indirect quotations.

\modelname extracts quotations by utilizing dependent relations between sequenced texts. 
The model contains three components, \ie \textit{Text Encoder}, \textit{Enhanced Cell}, and  \textit{Label Assigner}. 
Given a piece of text, the encoder encodes the instance and outputs the encoded hidden vectors. 
We design the Enhanced Cell module to study semantic representations of variable-length components with the utilization of contextual information. Specifically, the enhanced cell
(i) uses a composer layer to enhance the input with the former labels (which are predicted by the former cells), the former words, the current word, and the latter words encoded by the encoder; and
(ii) uses a gate layer and an attention layer to control and attend the corresponding input when predicting the label of the current word, at the level of element and vector respectively.
Experimental results on two public datasets (\ie \polnear and \riqua) and one proprietary dataset (\ie \datasetzh) show that our \modelname achieves state-of-the-art performance on complicated quotation extraction.

\section{Related Work}

At first glance, quotation detection is a kind of ``triplet'' extraction, making the task similar to another two tasks, open information extraction~\cite{angeli-etal-2015-leveraging,DBLP:conf/emnlp/GashteovskiGC17} and semantic role labeling~\cite{DBLP:conf/semweb/ExnerN11}. However, these three tasks have different focuses. 
Arguments extracted by semantic role labeling are event-related factors. OpenIE aims to output a structured representation of an instance in the form of binary or n-ary tuples, each of which consists of a predicate and several arguments. The extracted text spans in both tasks are typically short and less complicated, compared to the \textit{content} in quotations. Because \textit{content} extraction is the key challenge in quotation extraction, we will not further elaborate on semantic role labeling and OpenIE. 
Prior work on quotation extraction can be grouped into rule-based and sequence labeling methods.

\subsection{Rule-based Methods}
Extracting indirect quotations without clear boundaries is a challenging task, so early studies focus on rule-based methods to extract direct quotations~\cite{pouliquen2007automatic,krestel-etal-2008-minding,Elson2010AutomaticAO}. In fact, rule-based methods perform well for marked texts, especially for direct quotations.

Pattern matching is a popular method in early studies. 
~\citet{pouliquen2007automatic,Elson2010AutomaticAO} identify  \textit{content}, \textit{cue} and \textit{source} by  known quote-marks, pre-defined vocabulary, and rules of pattern recognition. The difference is that \citet{Elson2010AutomaticAO} add machine learning methods to the quote attribution judgment so that they can process complex text. \citet{DBLP:conf/emnlp/OKeefePCKH12} use regular expressions to recognize quote-marks to extract components, then use sequence labeling to recognize quotation triplets.

Hand-built grammar is another popular rule-based method.
~\citet{krestel-etal-2008-minding} design a system by combining common verbs corresponding to \textit{cue} and hand-built grammar to detect constructions that match six general lexical patterns. PICTOR~\cite{Schneider2010VisualizingTQ} utilizes context-free grammar to extract  components of quotations.

\subsection{Sequence Labeling Methods}

Due to the development of deep learning, sequence-labeling-based approaches have attracted attention~\cite{pareti-etal-2013-automatically,DBLP:conf/www/LeeCCH20}. 
To identify the beginning of a quotation, \citet{fernandes-etal-2011-quotation} use sequence labeling with  features including  part-of-speech and entity features generated by a guided transformation learning algorithm. Then they use regular expressions to recognize the \textit{content} within quotations.
\citet{pareti-etal-2013-automatically} follow a similar idea but use CRF to decode the label. \citet{DBLP:conf/www/LeeCCH20} further use BERT to encode the text and CRF to decode the label on a non-public Chinese news dataset. However, these models cannot well handle quotations with complicated structures.

\begin{figure*}
    \centering
    \includegraphics[scale=0.13]{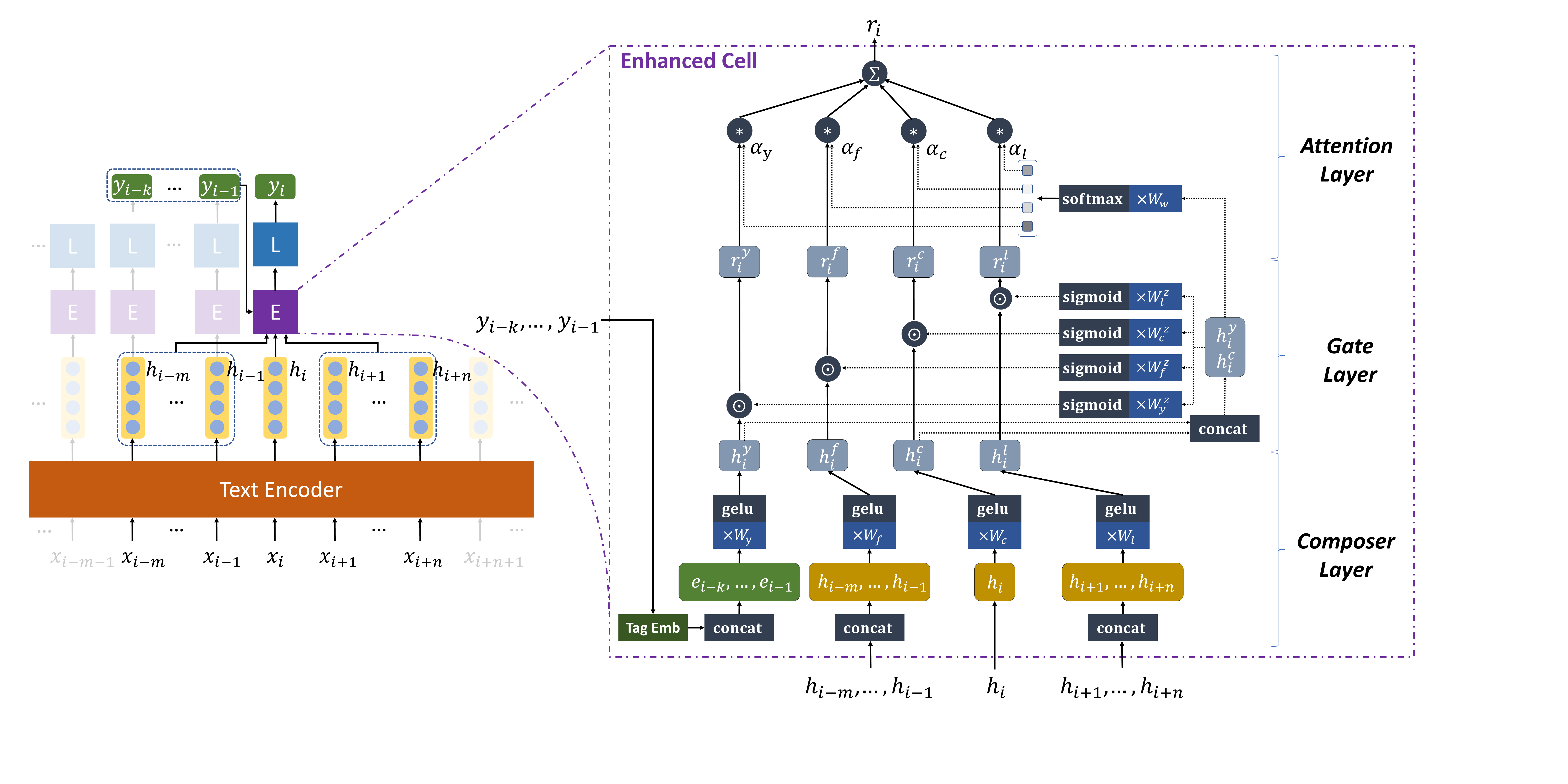}
    \caption{
    The architecture of \modelname. Enhanced Cell is detailed on the right-hand side. (best viewed in color) 
    }
    \label{fig:model}
\end{figure*}

\section{\modelname Model}
\label{sec:model}

Figure~\ref{fig:model} depicts the architecture of \modelname.
It consists of three modules:  \textit{Text Encoder},  \textit{Enhanced Cell}, and  \textit{Label Assigner}. 
Text encoder is used to encode the input text to get  hidden representations. Then, the enhanced cell is capable of building a representation considering the trait of quotations including variable-length and complicated-structured components. Last, the label assigner is to assign labels ``B-source'', ``B-cue'', ``B-content'', ``I-source'', ``I-cue'', ``I-content'' and ``O'', with BIO scheme.

\subsection{Text Encoder}

\modelname is generic and can be realized by popular encoders such as LSTM~\cite{DBLP:journals/neco/HochreiterS97}, CNN~\cite{kim-2014-convolutional}, Recursive Neural Network~\cite{DBLP:conf/icml/SocherLNM11}, and BERT~\cite{devlin-etal-2019-bert}. Unless otherwise specified, \modelname denotes the model using BERT~\cite{DBLP:conf/naacl/DevlinCLT19} as the encoder.

Given input text,  hidden states of words are formulated by: 
\begin{equation}\notag
    \{h_1, h_2, \dots, h_N\} =\text{Encoder}(\{x_1, x_2, \dots, x_N\}),
\end{equation}
where, $x_i$ is the $i$-th word of input, and \text{Encoder} denotes the Text Encoder. 
The hidden state $h_i$ denotes the representation of $i$-th word $x_i$ while encoding the preceding contexts of the position. 

\subsection{Enhanced Cell}

As aforementioned, the challenge of quotation extraction is to extract the complicated-structured components with variable lengths. To this end, we design the enhanced cell with composer layer, gate layer, and attention layer, to study the semantic representations of variable-length components. At the same time, we also try to utilize contextual information and predicted labels. 

Shown in Figure~\ref{fig:model}, the composer is used to reformat the input information to include the former labels $y_{i-k}, \dots, y_{i-1}$, the former hidden states $h_{i-m}, \dots, h_{i-1}$, the current state $h_i$, and the latter states $h_{i+1}, \dots, h_{i+n}$. In this way, our model is able to consider a long span with different structures in a more coherent manner on top of encoded word representations. 
In general, the influence of different inflow information is different. To this end, 
we use a gate mechanism to control each element of input representations, and an attention mechanism to weigh the input representations at the vector level. Through the two mechanisms, we get a refined representation so that we could hold the complicated-structured and variable-length components of quotations.
Next, we detail the workflow of the enhanced cell.

\paratitle{Composer Layer.} The composer contains a label embedding unit and a linear unit to reformat the inflow information:  the former labels \{$y_{i-k}, \dots, y_{i-1}\}$, the former hidden states $\{h_{i-m}, \dots, h_{i-1}\}$ of previous $m$ words, the current state $h_i$ of the current word $x_i$, and the latter states $\{h_{i+1}, \dots, h_{i+n}\}$ of latter $n$ words.

First, the enhanced cell contains a label embedding unit, which is able to select the embedding of the given label, formulated by:
\begin{equation}
    e_i = \text{Emb}(y_i),
\end{equation}
where $\text{Emb}$ denotes the mentioned label embedding unit. The predicted label of word $i$ is $y_i$ and the embedding of $y_i$ is $e_i$. Taking the former $k$ predicted labels into consideration, we get the former labels' representations $[e_{i-k}, \dots, e_{i-1}]$ by concatenation, which is shown as a rectangle in green background, in the Enhanced Cell in Figure~\ref{fig:model}.

Intuitively, contextual information is important for us to predict the label of the current input word.  We take the following context through simple but effective linear layers: the former predicted $k$ labels, the former $m$ words, the current word $i$, and the latter $n$ words.
\begin{align}
    h_{i}^{y} &= \gelu([e_{i-k}, \dots, e_{i-1}] W_y + b_y) \\
    h_{i}^{f} &= \gelu([h_{i-m}, \dots, h_{i-1}] W_f + b_f) \\
    h_{i}^{c} &= \gelu(h_{i} W_c + b_c) \\
    h_{i}^{l} &= \gelu([h_{i+1}, \dots, h_{i+n}] W_l + b_l)
\end{align}
In the above formulation, the hidden states $\{h_{i-m}, \dots, {h_i}, \dots, h_{i+n}\}$ and label embeddings $\{e_{i-k}, \dots, e_{i-1}\}$ are the input. $W_y, W_f, W_c, W_l$ and $b_y, b_f, b_c, b_l$ are the parameters of the linear layers. Here, we adopt $\gelu$ as the active function. 
$h_i^y, h_i^f, h_i^c, h_i^l$ denote the farther hidden states of the former labels, the former words, the current word and the latter words, respectively.

\paratitle{Gate Layer.}
The influence of different contexts is different. Hence, we use a gate mechanism to control the inflow hidden states at the element level. 
Inspired by~\citet{DBLP:journals/neco/HochreiterS97}, we design a gate layer in the enhanced cell:
\begin{align}
    r_{i}^{y} &= h_{i}^{y} \odot \sigmoid([h_{i}^{y}, h_{i}^{c}] W_{y}^{z} + b_{y}^{z}) \\
    r_{i}^{f} &= h_{i}^{f} \odot \sigmoid([h_{i}^{y}, h_{i}^{c}] W_{f}^{z} + b_{f}^{z}) \\
    r_{i}^{c} &= h_{i}^{c} \odot \sigmoid([h_{i}^{y}, h_{i}^{c}] W_{c}^{z} + b_{c}^{z}) \\
    r_{i}^{l} &= h_{i}^{l} \odot \sigmoid([h_{i}^{y}, h_{i}^{c}] W_{l}^{z} + b_{l}^{z})
\end{align}
In the above formulation, $r_{i}^{y}$, $r_{i}^{f}$, $r_{i}^{c}$, and $r_{i}^{l}$ denote the adjusted states of the former labels, the former words, the current word, and the latter word representation, respectively.
The operation $\odot$ denotes element-wise product.
$W_{y}^{z}$, $W_{f}^{z}$, $W_{c}^{z}$, $W_{l}^{z}$, and $b_{y}^{z}$, $b_{f}^{z}$, $b_{c}^{z}$, $b_{l}^{z}$ are the parameters. We use $\sigmoid$ to adjust each element of the inflow representations.

\paratitle{Attention Layer.} 
Inspired by~\citet{wang-etal-2016-attention,yang-etal-2016-hierarchical,DBLP:conf/www/WangSH0Z18,DBLP:conf/cikm/LinWXLB19,DBLP:conf/kdd/MengWMLLZ22}, we use an attention mechanism to attend the important part of $r_{i}^{y}$, $r_{i}^{f}$, $r_{i}^{c}$, and $r_{i}^{l}$.
Since our target is to predict the label of the current word, we use the concatenation of $h_{i}^{y}$ and $h_{i}^{c}$ to attend the four vectors by
\begin{equation}
    \alpha_y, \alpha_f, \alpha_c, \alpha_l = \softmax([h_{i}^{y}, h_{i}^{c}] W_{w} + b_w),
\end{equation}
where $\alpha_y, \alpha_f, \alpha_c$, and $\alpha_l$ are the weights for $r_{i}^{y}$, $r_{i}^{f}$, $r_{i}^{c}$, and $r_{i}^{l}$ respectively. $W_w$ and $b_w$ are the parameters. In the attention layer,  $\softmax$ function is used to calculate weights. Then, the current word representation $r_i$ is obtained via:
\begin{equation}
    r_i = \alpha_y r_{i}^{y} + \alpha_f r_{i}^{f} + \alpha_c r_{i}^{c} + \alpha_l r_{i}^{l}
\end{equation}

To summarize, the Enhanced Cell uses the gate and attention layers with contextual information (\ie former labels, former words, current word, and latter words) to handle complicated-structured components with variable lengths. 
Specifically, to sense continuous span, we use attention layer by attending contextual information at the vector~(macro) level, by using former labels, and the former, current, and latter word(s). Thus, the model avoids undesirable interruption within an instance.
We also use the gate layer to control contextual information at the element~(micro) level, especially former labels.
Further, thanks to the ability of fine control, the gate layer is capable of avoiding illegal patterns, \eg ``O'' followed by ``I-*''.

\subsection{Label Assigner}

After getting the hidden representation of the current word, we use label assigner module to compute a probability distribution of the current label.

Briefly speaking, in label assigner, we use $\softmax$ classifier to calculate the distribution $\mathcal{P}_i$ of the current word $i$. Then $\argmax$ is used to assign a label of the current word. The two operations can be formulated as
\begin{align}
    \mathcal{P}_i &= \softmax(r_i W_p + b_p), \\
    y_i &= \argmax(\mathcal{P}_i),
\end{align}
where $W_p$ and $b_p$ are the parameters.

\subsection{Training Objective}

The proposed \modelname model could be trained in an end-to-end way by backpropagation.
We adopt the cross-entropy objective function that has been used in many studies~\cite{tang-etal-2015-document,wang-etal-2016-attention,DBLP:conf/www/WangSHZ19}.

\paratitle{Sequence Labeling Objective.} Similar to sequence labeling tasks, we evaluate the label of all words for each given training instance. Recall that our objective is to predict the label of each word in the given instance. The unregularized objective $L$ can be formulated as cross-entropy loss:
\begin{equation}
    L(\theta) = -\sum_{i}\sum_j l_i^j \log(\mathcal{P}_i^j)
\end{equation}
For a given training instance, $l_i^j$ is the ground truth of label $j$ for word $i$. Correspondingly, $\mathcal{P}_i^j$ is the probability of label $j$ for word $i$. $\theta$ is the parameter set.

\section{Experiment}
\label{sec:exp}

We now evaluate the proposed \modelname on two public datasets (\ie \polnear  and \riqua), and one proprietary dataset (\ie \datasetzh) against baselines. The implementation details and parameter settings are presented in Appendix~\ref{sec:app:imp}. On all datasets, we train the model with the training set, tune hyperparameters on the validation set, and report performance on the test set.

\subsection{Datasets}

\paratitle{\polnear.} Political News Attribution Relations Corpus~(\polnear)~\cite{DBLP:conf/lrec/NewellMR18} is a corpus of news articles in English, on political candidates during US Presidential Election in November 2016.
\polnear annotations are univocal, meaning that each word has only one label (\textit{source}, \textit{cue}, \textit{content}, or none).
The average number of tokens is $46$. 

\paratitle{\riqua.} RIch QUotation Annotations~(\riqua)~\cite{DBLP:conf/lrec/PapayP20}  provides quotations, including interpersonal structure (speakers and addressees) for English literary. This corpus comprises 11 works of 19th-century literature that are manually annotated for direct and indirect quotations. 
Each instance, typically a sentence, is annotated with its \textit{source}, \textit{cue}, and \textit{content}. The average number of tokens in this corpus is $129$, longer than \polnear.

\paratitle{\datasetzh.} Chinese Political Discourse~(\datasetzh) contains politics and economics news collected from mainstream online media of China including Xinhua Net\footnote{\url{http://news.cn/}}.
The news are in Chinese and the average length of input is $69$ tokens, longer than \polnear but shorter than \riqua.

Table~\ref{tab:stat} presents the statistics of the three datasets. We observe that the numbers of instances of \polnear and \datasetzh are at the order of $10k$, and the \riqua is at $1k$. The length of \textit{source} and \textit{cue} is less than $5$ tokens. The length of \textit{content} is greater than $10$, even $40$ tokens. Note that for all three datasets, the length of \textit{content} is much longer than \textit{source} and \textit{cue}.

\subsection{Compared Methods}

\input{tables/stat}

\input{tables/main_result}

To provide a comprehensive  evaluation, we experiment on both deep learning (\ie CNN, GRU, (Bi)LSTM, BERT, and BERT-CRF), and traditional methods (\ie Rule and CRF).

\paratitle{Rule.} \citet{okeefe-etal-2012-sequence} uses rules including entity dictionary, reported speech verbs, and special flag characters to extract components of quotations.

\paratitle{CoreNLP.} CoreNLP~\cite{vu-etal-2018-vncorenlp} contains quote extraction pipeline which deterministically picks out \textit{source} and \textit{content} from a text while ignoring \textit{cue}.

\paratitle{CRF.} \citet{DBLP:conf/icml/LaffertyMP01} present CRF to  label sequence by building probabilistic models. 
    
\paratitle{CNN.} CNN~\cite{lecun1995convolutional}, a simple and parallelized model, can be independently adopted for sequence labeling tasks~\cite{xu-etal-2018-double}. 

\paratitle{(Bi)LSTM.} LSTM~\cite{DBLP:journals/neco/HochreiterS97} is able to exhibit dynamic temporal behavior due to its well-designed structure. We use it and its variants, \ie Bidirectional LSTM~(BiLSTM).
    
\paratitle{GRU.} GRU is a slightly more dramatic variation of LSTM~\cite{cho-etal-2014-learning}.
    
\paratitle{BERT(-CRF).} BERT is designed to pre-train deep bidirectional representations from unlabeled text by jointly conditioning on both left and right contexts~\cite{devlin-etal-2019-bert}.

\subsection{Evaluation Metrics}
\label{sec:evaluation}

The components of quotations are variable-length and complicated. As a result, it requires more specific metrics. To this end, we evaluate the performance of models using our proposed ``Jaccard'', in addition to ``Exact Match'' and ``Begin Match''.

\paratitle{Exact Match.} To measure the overall prediction at the instance level, we propose  $\text{Exact Match}$ index to quantify whether the multi-label prediction exactly matches the annotation.
In the experiments,  we use accuracy, precision, recall, and $F1$ to evaluate the exact match performance.

\paratitle{Begin Match.} Exact match is harsh, especially long text span. Generally, the length of \textit{source} and \textit{cue} is short while the \textit{content} is much longer. As a result, exact match is hard for \textit{content}. To this end, we use begin match to evaluate only the beginning location for text span matching \cite{LeeS19}.

\paratitle{Jaccard.} For text span matching, an important index is a ratio of the overlapping span over the total span. Thus we use ``Jaccard'' index to evaluate the performance of model in this aspect. 
Given the groundtruth text span $\mathcal{T}_g$ and its predicted text span $\mathcal T_p$, we can calculate the Jaccard index $J$  through
\begin{equation}
    J =\frac{|\mathcal T_p \cap \mathcal T_g|}{|\mathcal T_p \cup \mathcal T_g|}.
\end{equation}

\subsection{Main Results}

Table~\ref{tab:main_result} lists the $F1$ and $J$(accard) performance on the three datasets. In this table, the best results are in boldface and the second-best are underlined. 
We report results by exact match, begin match, and Jaccard, of all models for the three components of quotations. Here, $F1$-E. and $F1$-B. refer to the  $F1$ based on exact match and begin match, respectively. 
The precision, recall and accuracy are shown in the page\footnote{\url{https://thuwyq.github.io/docs/cofenet-detail-exp.pdf}} due to space limitation. 
Our \modelname model is listed in the last row of each dataset.

\input{tables/compare_crf}

Table~\ref{tab:main_result} shows that our \modelname performs the best against all baselines. BERT achieves the second-best, followed by other deep-learning-based models. Note that due to the settled human-written rules, the performance of Rule and CoreNLP is not stable. For \textit{source} and \textit{cue}, on \datasetzh, the performance is good due to more comprehensive rules. However, the rules on the other two datasets do not fit the domain well.
As a comparison, \textit{content} is on the opposite side.
For \textit{content}, the precision and recall of CoreNLP are $97.2$ and $47.5$ on \riqua dataset, which is better than \polnear. \datasetzh dataset shows the worst performance. 
This is because CoreNLP uses quote marks to extract quotations. The number of direct quotations (\ie quoted \textit{content}) on \polnear and \riqua is large, while the \datasetzh is small.
This shows that the rule-based methods cannot effectively identify indirect quotations.

The level of difficulty in extracting \textit{source}, \textit{cue}, and \textit{content} is different. As a result, the performances of \textit{source} and \textit{cue} are better than the difficult \textit{content}. This is expected because \textit{content} is longer and complex in semantics. For example, the \textit{content} may contain another \textit{source}, \textit{cue} and \textit{content}. We design gate and attention mechanisms to fit those so that our model performs well.

\subsection{Comparison with CRF and BERT}

\paratitle{Comparison with CRF.} CRF is a popular approach to handle sequence labeling problems, \eg NER~\cite{ritter-etal-2011-named,DBLP:conf/nlpcc/DongZZHD16}. 
We compare \modelname with CRF by changing the encoder, \ie \textit{LSTM w. Cofe} denotes the Cofe using LSTM as text encoder. Recall that \modelname specifically refers to the model using BERT as encoder, marked as \textit{BERT w. Cofe} in Table~\ref{tab:compare_crf}.
To make the comparison comprehensively and deeply, our comparisons  between CRF and Cofe are based on various mainstream models including CNN, GRU, LSTM, BiLSTM, and BERT.

Table~\ref{tab:compare_crf} details the comparison results on \polnear, and the results of the other two datasets are reported in the page\footnote{\url{https://thuwyq.github.io/docs/cofenet-detail-exp.pdf}}.
(i) Results show that both Cofe and CRF perform better than basic models, and Cofe-based models perform better than CRF-based models.
The comparison results suggest that our model architecture fits well with  dependent sequence labeling tasks. As designed, the enhanced cell is capable of building the dependency relations of labels. 
(ii) Another interesting observation from the results is that if the basic model (\eg GRU) is simple, a larger improvement is achieved. On the contrary, the improvement over BERT is relatively small. It makes sense because the improvement is harder when the performance is already at a very high level. 
(iii) We also note that \modelname performs better than CRF on all components of quotations.

\paratitle{Comparison with BERT.} 
BERT based models are strong baselines for many tasks, particularly when there are clear patterns.
The performance of models could be improved if we adopt a dependent encoding method based on BERT. To this end, based on BERT, we use decoders including CNN, LSTM, BiLSTM, BiLSTM+CRF in addition to CRF. The bottom area of Table~\ref{tab:compare_crf} shows the results. 
Results show that the improvements of decoders including CNN, LSTM and BiLSTM are not significant than BiLSTM+CRF. Despite this, our \modelname performs best.
When meeting simple text span (\eg Cue), the improvement of our proposed \modelname is relatively small (1.3 point improvement, F1-Exact Match, on the Cue of PolNeAR dataset). When it comes to complex text span (\eg Content), our model shows large improvement over BERT model (4.0 points improvement, F1-Exact Match, on the Content of PolNeAR dataset).

From the comparisons, we demonstrate that our proposed \modelname achieves the state-of-the-art performance on quotation extraction. To reveal the essence of \modelname, we show the transition matrix of labels,  the analysis on attention mechanism, and the ablation study in the next sections.

\subsection{Label Transition Matrix}

\begin{figure}
    \subfigure[The transition matrix of groundtruth]{
    \hspace{-2ex}
    \includegraphics[scale=0.52]{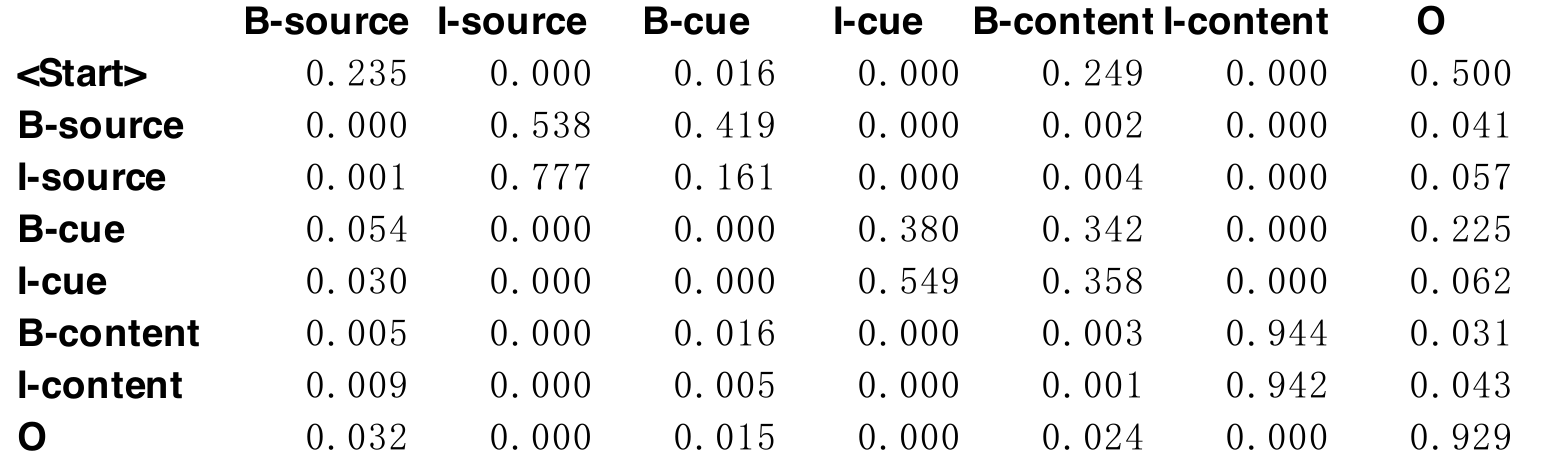}
    \label{fig:lt-ground}
    }
    \subfigure[The margin between groundtruth and \modelname]{
    \hspace{-2ex}
    \includegraphics[scale=0.52]{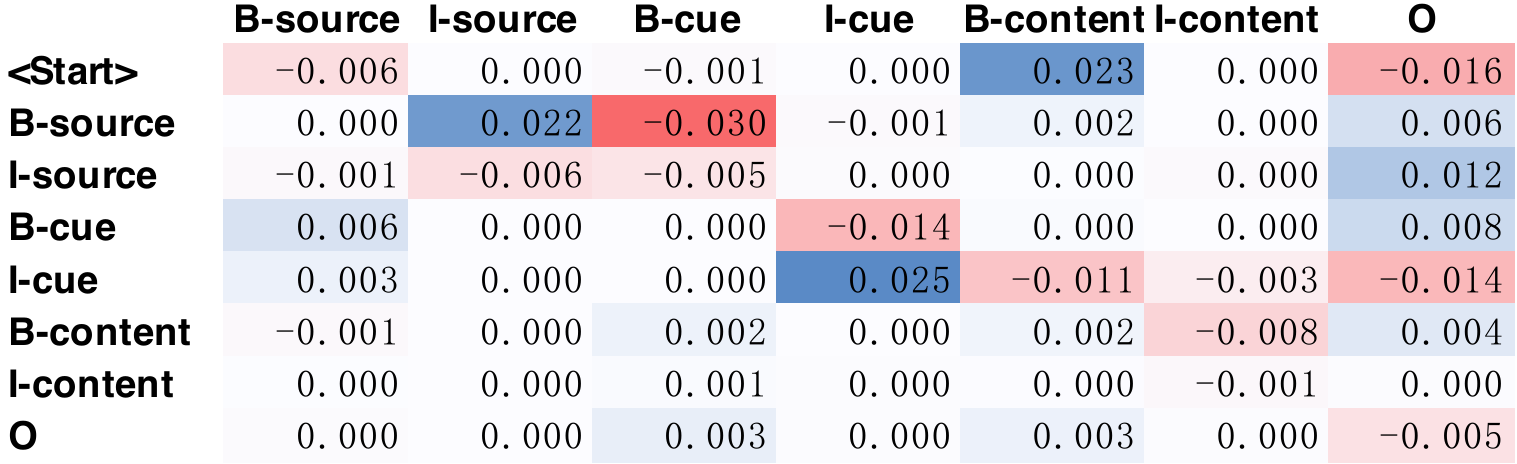}
    \label{fig:lt-margin}
    }
    \caption{The transition matrix and the margin of groundtruth and our model on \polnear.}
    \label{fig:label-transfer}
\end{figure}

The probability transition matrix of labels reflects the particular features of \textit{source}, \textit{cue} and \textit{content}. Thus
we can use them to reveal the transition mechanism of labels. To this end, we calculate the label transition matrix of groundtruth, and the margin between groundtruth and \modelname.
Figure~\ref{fig:label-transfer} depicts the detail on \polnear. In all subfigures, the column denotes the previous label and the row represents the current label. The value of Figure~\ref{fig:lt-ground} denotes the transition probability of true labels, and the value of Figure~\ref{fig:lt-margin} is the margin between the true and the predicted.
As the word saying, ``\textlangle{Start}\textrangle'' denotes the location before the first word, ``B-'' and ``I-'' denote the beginning and the inside of the \textit{source}, \textit{cue} and \textit{content}, respectively. ``O'' refers to the other words.

The transition matrix of groundtruth shown in Figure~\ref{fig:lt-ground} reveals the statistics of the \polnear dataset. Recall that the key for quotation extraction is the recognition of the ``Begin''. Hence, the margin of ``Begin'' is the compass for evaluating the performance. We find that the maximum absolute margin of ``Begin'' is $-0.03$, when the precious label is ``B-source'' and the current label is ``B-cue''. This is because the length of \textit{source} is short, and  \textit{cue} word often follows  \textit{source} word closely. This proves that our model performs well even in difficult situations.

For BIO labeling scheme, the ``I-source/cue/con-tent'' exists except the corresponding ``B-*'' exists. As a result, the transition value of ``I-'' could show the recognition ability of the model for those patterns. Also, Figure~\ref{fig:lt-margin} shows almost all margins of those values are zeros. This reveals that our model could study those key patterns well.

\subsection{Analysis on Attention Mechanism}
\label{sec:exp:weight}

\begin{figure}
    \centering
    \hspace{-1mm}\includegraphics[scale=0.38]{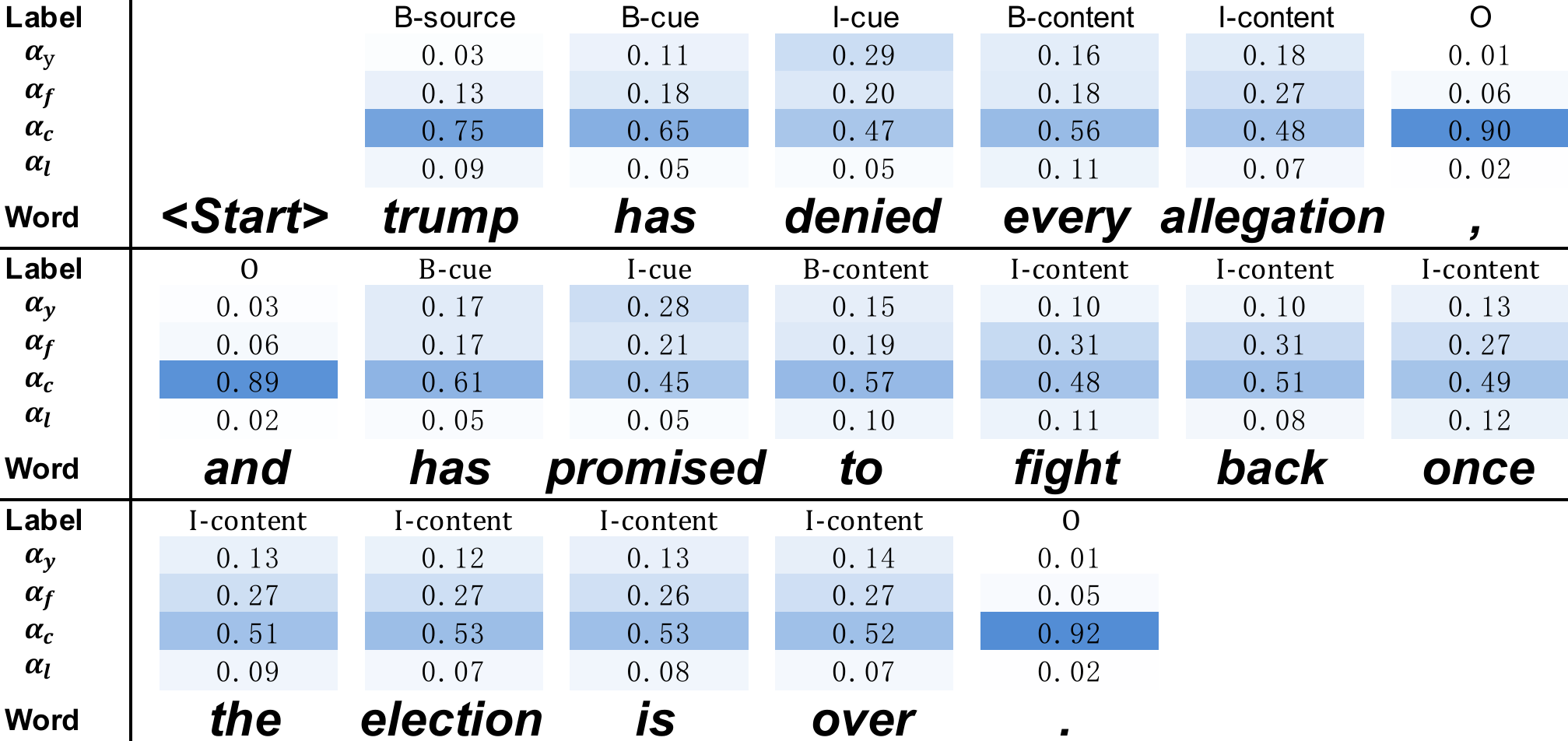}
    \caption{The attention weights of one test data from \polnear.}
    \label{fig:case}
\end{figure}

In our design, the utilization of inflow information (\eg  former labels,  previous words, current word, and  latter words) is the key for quotation extraction. 
Figure~\ref{fig:case} shows the weights from the attention layer of one test instance in \polnear. To avoid the bias of a single case, we do a global prediction for all texts in the test dataset of \polnear attached in Appendix~\ref{sec:app:attention}.
(i) The current word information has the largest weight, as expected. For the prediction of ``I-source/cue/content'', the former labels and former words information are the most important roles after the current word.
It indicates that our model is capable of utilizing the former labels and sequence information as we designed. 
(ii) Another interesting observation is that the weights of the latter words' information for predicting ``B/I-content'' are about $0.1$, which are greater than the other weights in $\alpha_l$.
As we mentioned before, the length of \textit{content} is longer than \textit{source} and \textit{cue}, so the utilization of latter information improves the performance of long-span extraction more efficiently.

\input{tables/ablation}

\subsection{Ablation Study}

The \modelname model  uses gate mechanism $g.m.$ and attention mechanism $a.m.$ (see Section~\ref{sec:model}) to utilize information including former labels $f.l.$, former words $f.w.$, current word $c.w.$, and latter words $l.w.$. 
To study the effect of the two mechanisms and on the four information sources, we conduct ablation experiments on \polnear dataset.

Table~\ref{tab:ablation} reports the results of this ablation study. (i) As expected, all mechanisms and information are useful for quotation extraction. 
For \textit{content}, the Jaccard performance degrades at least $1.0$ points after removing mechanisms or input information, which is similar to \textit{source} and \textit{cue}. As a comparison, the performance drop on $F1$-E. and $F1$-B. is significantly less than $J$. It is because the structure of \textit{source} and \textit{cue} is simpler than \textit{content}.
This phenomenon shows our \modelname is particularly suitable for extracting quotations with long and complicated structures.
(ii) When removing attention, larger drops on exact match are observed than removing gate. It reveals that attention is effective for begin match while gate prefers exact match.
(iii) Further, we explore the performance of inflow information. The ``w.o. f.w.'' on Table~\ref{tab:ablation} shows that the former words' information is not so important for the prediction of \textit{cue} because the \textit{cue} is the shortest of all three components.
The former label and the current word, the latter words are important for all of the components. It proves that the latter words' information is key for the recognition of \textit{content}. This fits with our observations in Section~\ref{sec:exp:weight}. 

\section{Conclusion and Future Work}
In this study, we design the \modelname model for quotation extraction with variable-length span and complicated structure. The key idea of \modelname model is to use gate and attention mechanisms to control the important information including  former labels,  former words,  current word and  latter words at the element and vector levels. Experiments show that the proposed model achieves the state-of-the-art performance on two public datasets \polnear and \riqua and one proprietary dataset \datasetzh. 

For quotation analysis, the extraction of quotation components is the first step. In our study, we split a long text into short texts to ensure that one instance contains one \textit{source}, one \textit{cue} and one \textit{content}. Thus the recognition of quotation triplets from long text (\eg across instance) is one important future work. Another important direction is to go deep into the nesting phenomenon, which makes the recognition harder.

\section*{Acknowledgments}

This work was supported by the National Key R\&D Program of China (2020AAA0105200) and the National Science Foundation of China (NSFC No. 62106249, 61902382, 61972381).

\bibliography{anthology,acl}

\clearpage
\appendix

\section*{Appendix}
\label{sec:app}

\section{Implementation Details}
\label{sec:app:imp}

We list the implementation details of \modelname.

\input{tables/model-setting}

Table~\ref{tab:hyper-param} lists the same settings for the two public datasets (\ie \polnear and \riqua) and our proprietary dataset (\ie \datasetzh). The learning rate for model parameters except BERT are $1e-3$, and $5e-5$ for BERT.
We use typical $12$-layers BERT (known as \textit{bert-base-uncased}~{\footnote{https://s3.amazonaws.com/models.huggingface.co/bert/bert-base-uncased-pytorch\_model.bin}}) as a basic encoder for the two English datasets. For the Chinese dataset \datasetzh, we use \textit{bert-base-chinese}~\footnote{https://s3.amazonaws.com/models.huggingface.co/bert/bert-base-chinese-pytorch\_model.bin}.
The middle part of Table~\ref{tab:hyper-param} shows the important hyperparameters of BERT.
There are other hyperparamters for \modelname except BERT related. The hidden sizes of word representation and label embedding are 100. The number of former labels, former words, and latter words is $1$, $3$, and $3$, respectively. 
The different hyperparameter for \modelname is the batch size due to the GPU memory limitation. During training, we set the batch sizes for \polnear, \riqua and \datasetzh to $15$, $15$ and $16$, respectively.

\begin{figure}[ht]
    \subfigure[The weight $\alpha_y$ for former labels $r_{i}^{y}$]{
    \hspace{-2ex}
    \includegraphics[scale=0.52]{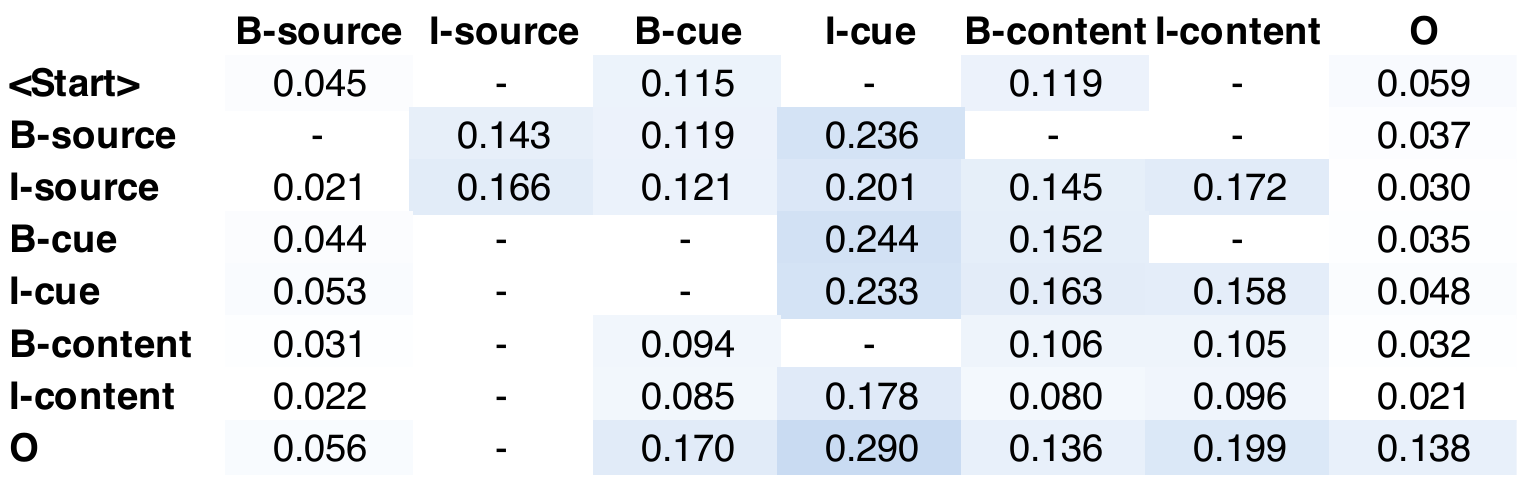}
    \label{fig:weight-fl}
    }
    \subfigure[The weight $\alpha_f$ for former words $r_{i}^{f}$]{
    \hspace{-2ex}
    \includegraphics[scale=0.52]{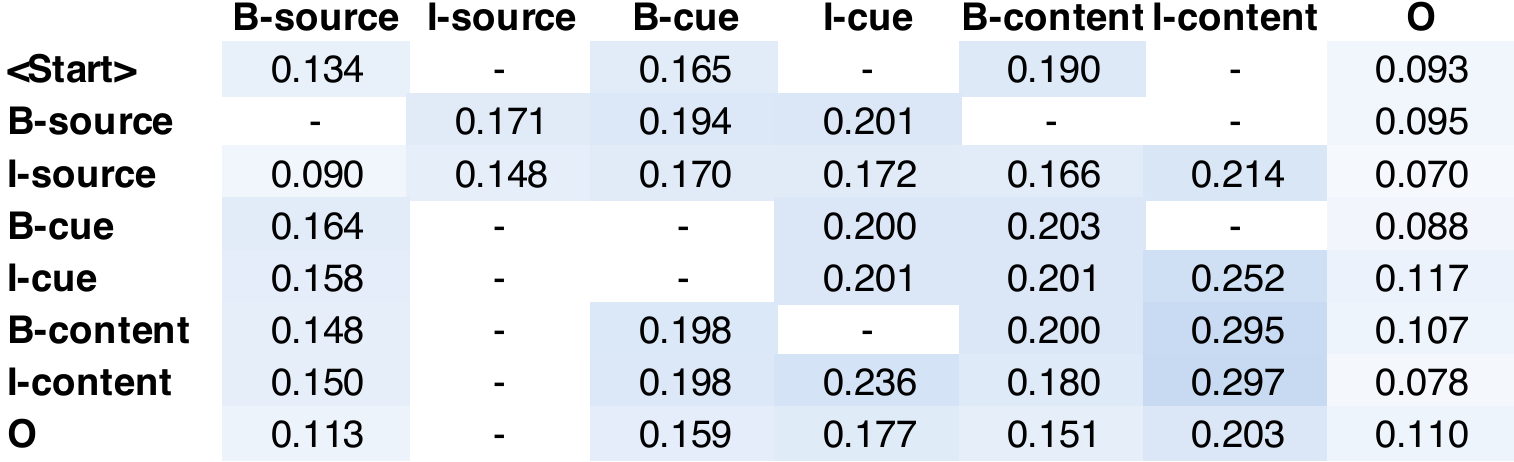}
    \label{fig:weight-fw}
    }
    \subfigure[The weight $\alpha_c$ for current word $r_{i}^{c}$]{
    \hspace{-2ex}
    \includegraphics[scale=0.52]{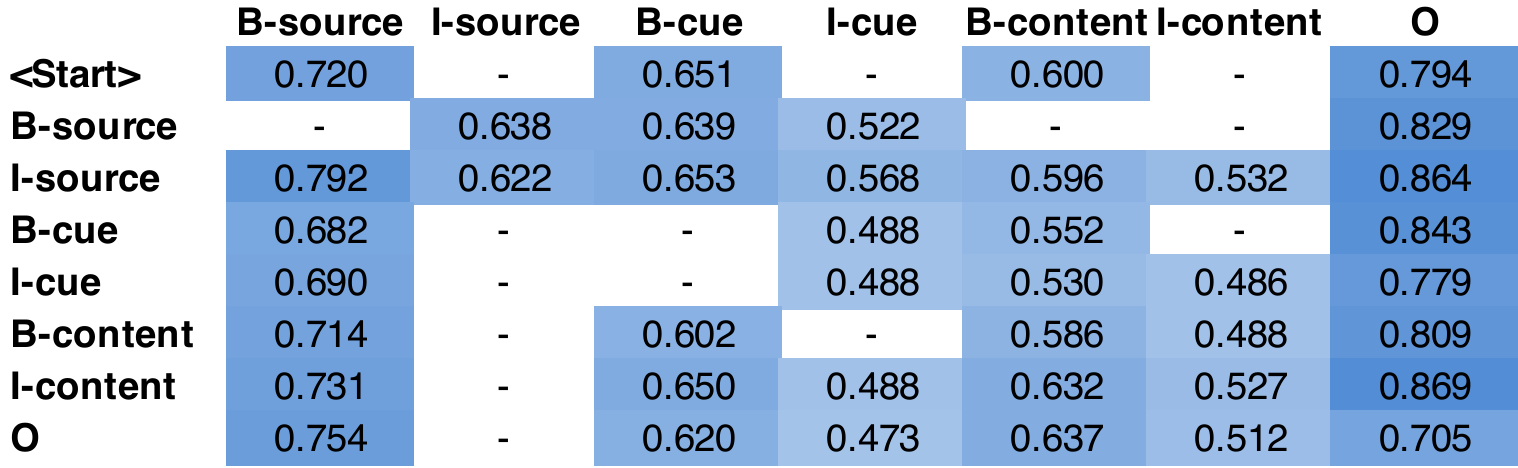}
    \label{fig:weight-cw}
    }
    \subfigure[The weight $\alpha_l$ for latter words $r_{i}^{l}$]{
    \hspace{-2ex}
    \includegraphics[scale=0.52]{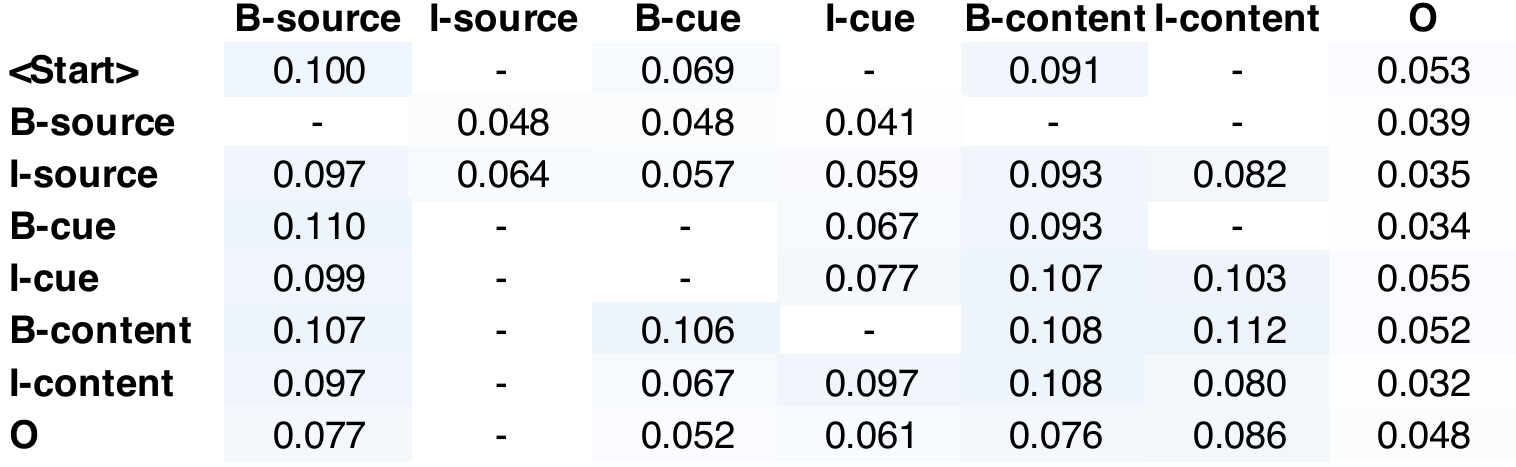}
    \label{fig:weight-lw}
    }
    \vspace{1ex}
    \caption{The weights for hidden states on \polnear.}
    \label{fig:weight}
    \vspace{-3ex}
\end{figure}

We use Adam~\cite{DBLP:journals/corr/KingmaB14} as our optimization method. \modelname is implemented on Pytorch~(version 1.2.0). NLTK is used to segment text.
For BERT model, we invoke the pytorch-transformers package~(version 1.2.0). 
To ensure the reliability of experimental results, we use the same transformer package with the same initialization parameters in BERT, BERT-CRF and \modelname.

\section{Global Analysis on Attention Mechanism}
\label{sec:app:attention}

In our design, the utilization of inflow information is the key for quotation extraction. Recall that the information includes the former labels, the previous words, the current word and the latter words. Hence, we use the attention to reveal the operating principle of the model.
Figure~\ref{fig:case} has shown the weights from the attention layer of one individual case from test set of \polnear dataset. To avoid the bias of a single case, we do a global prediction for all texts in test set of \polnear shown in Figure~\ref{fig:weight}. The observations from Figure~\ref{fig:weight} are similar to that reported in Section~\ref{sec:exp:weight}, so we will not repeat them.

\end{document}

%% file: tables/stat.tex
\begin{table}
  \centering
  \caption{The statistics of three datasets. ``Ave. len.'' refers to ``Average length''.}
  \small\setlength{\tabcolsep}{1.7mm}{
    \hspace{-2.5mm}\begin{tabular}{l|rrr|rrr}
    \toprule
    \multicolumn{1}{l|}{\multirow{2}[4]{*}{Dataset}} & \multicolumn{3}{c|}{Number of sentences} & \multicolumn{3}{c}{Ave. len. in tokens} \\
\cmidrule{2-7}          & \multicolumn{1}{p{0.5cm}}{Train} & \multicolumn{1}{p{0.5cm}}{Valid} & \multicolumn{1}{p{0.5cm}|}{Test} & \multicolumn{1}{p{0.6cm}}{Source} & \multicolumn{1}{r}{Cue} & \multicolumn{1}{p{0.5cm}}{Content} \\
    \midrule
    \polnear & 17,397 & 1,925  & 1,814  & 3.27  & 1.88  & 14.49 \\
    \riqua & 1,604  & 208   & 105   & 1.38  & 1.08  & 20.65 \\
    \datasetzh & 10,754 & 1,344  & 1,345  & 3.08  & 1.80   & 43.47 \\
    \bottomrule
    \end{tabular}
    }
  \label{tab:stat}
\end{table}

%% file: tables/main_result.tex
\begin{table*}[ht]
  \centering
  \caption{The $F1$ and $J$(accard) of methods on \polnear, \riqua and \datasetzh datasets. The results marked with $^*$ are obtained by calling the CoreNLP toolkit package directly. }
  \small\setlength{\tabcolsep}{1mm}{
    \begin{tabular}{l|l|rrr|rrr|rrr}
    \toprule
    \multirow{2}[4]{*}{Dataset} & \multicolumn{1}{c|}{\multirow{2}[4]{*}{Model}} & \multicolumn{3}{c|}{Source} & \multicolumn{3}{c|}{Cue} & \multicolumn{3}{c}{Content} \\
\cmidrule{3-11}          &       & \multicolumn{1}{p{0.92cm}<{\centering}}{$F1$-E.} & \multicolumn{1}{p{0.92cm}<{\centering}}{$F1$-B.} & \multicolumn{1}{p{0.92cm}<{\centering}|}{$J$} & \multicolumn{1}{p{0.92cm}<{\centering}}{$F1$-E.} & \multicolumn{1}{p{0.92cm}<{\centering}}{$F1$-B.} & \multicolumn{1}{p{0.92cm}<{\centering}|}{$J$} & \multicolumn{1}{p{0.92cm}<{\centering}}{$F1$-E.} & \multicolumn{1}{p{0.92cm}<{\centering}}{$F1$-B.} & \multicolumn{1}{p{0.92cm}<{\centering}}{$J$} \\
    \midrule
    \multirow{10}[2]{*}{PolNeAR} 
          & Rule  & 10.7  & 13.0  & 8.8   & 22.8  & 25.3  & 14.4  & 5.6   & 10.5  & 6.1 \\
          & CoreNLP$^*$ & 13.9  & 21.3  & 11.1  & -   & -   & -   & 17.5  & 18.7  & 12.8 \\
          & CRF   & 50.6  & 56.2  & 42.1  & 53.4  & 63.3  & 44.1  & 28.6  & 50.9  & 42.3 \\
          & CNN              & 52.7  & 65.9  & 45.1  & 58.4  & 67.8  & 49.4  & 16.2  & 60.6  & 30.2 \\
          & GRU              & 46.5  & 58.2  & 36.7  & 59.1  & 68.1  & 48.8  & 51.3  & 65.0  & 51.3 \\
          & BiLSTM & 64.1  & 74.4  & 56.8  & 63.3  & 72.6  & 55.1  & 53.4  & 67.3  & 53.7 \\
          & BERT  & \underline{81.1}  & \underline{86.2}  & \underline{74.8}  & \underline{74.0}  & \underline{81.1}  & \underline{67.4}  & \underline{68.9}  & \underline{78.7}  & \underline{70.0} \\
          & \modelname & \textbf{83.2} & \textbf{87.1} & \textbf{76.4} & \textbf{75.3}  & \textbf{82.3} & \textbf{69.4} & \textbf{72.9} & \textbf{79.6} & \textbf{73.2} \\
    \hline
    \hline
    \multirow{10}[2]{*}{Riqua} 
          & Rule  & 16.8  & 16.8  & 11.2  & 36.5  & 36.5  & 22.3  & 0.0   & 2.4   & 2.4 \\
          & CoreNLP$^*$ & 22.8  & 22.8  & 17.9  & -   & -   & -   & 63.8  & 63.8  & 46.9 \\
          & CRF   & 46.9  & 51.0  & 32.9  & 59.6  & 65.7  & 46.6  & 42.7  & 85.9  & 62.2 \\
          & CNN              & 52.7  & 59.1  & 39.6  & 85.2  & 85.2  & 74.2  & 45.2  & 95.4  & 58.5 \\
          & GRU              & 55.8  & 62.9  & 43.4  & 77.1  & 77.1  & 62.8  & 92.5  & 95.2  & 89.6 \\
          & BiLSTM & 56.4  & 64.1  & 44.5  & 85.4  & 85.4  & 74.4  & 92.2  & 95.9  & 90.3 \\
          & BERT  & \underline{74.5}  & \underline{77.9}  & \underline{62.4}  & \underline{88.9}  & \underline{88.9}  & \underline{80.0}  & \underline{94.3}  & \underline{96.6}  & \underline{92.9} \\
          & \modelname & \textbf{81.8} & \textbf{84.3} & \textbf{72.6} & \textbf{89.2} & \textbf{89.2} & \textbf{80.4} & \textbf{94.4}  & \textbf{97.1} & \textbf{94.1} \\
    \hline
    \hline
    \multirow{10}[2]{*}{PoliticsZH} 
          & Rule  & 78.8  & 79.3  & 66.8  & 80.3  & 81.2  & 69.7  & 0.4   & 7.0   & 3.7 \\
          & CoreNLP$^*$ & 38.1  & 39.5  & 24.3  & -   & -   & -   & 0.2   & 2.2   & 4.3 \\
          & CRF   & 81.6  & 84.0  & 72.2  & 80.0  & 80.4  & 68.5  & 45.7  & 49.1  & 66.3 \\
          & CNN              & 82.5  & 87.8  & 76.5  & 81.4  & 83.6  & 72.1  & 35.0  & 74.5  & 46.7 \\
          & GRU              & 85.5  & 88.3  & 78.1  & 82.1  & 84.6  & 73.6  & 65.7  & 79.8  & 71.5 \\
          & BiLSTM & 87.5  & 91.3  & 83.3  & 86.2  & 88.6  & 79.9  & 70.3  & 81.8  & 74.9 \\
          & BERT  & \underline{92.6}  & \underline{93.7}  & \underline{88.2}  & \underline{89.5}  & \underline{90.8}  & \underline{84.0}  & \underline{73.7}  & \underline{83.6}  & \underline{84.4} \\
          & \modelname & \textbf{93.7} & \textbf{94.4} & \textbf{89.8} & \textbf{90.3} & \textbf{91.1} & \textbf{85.4} & \textbf{78.0} & \textbf{86.9} & \textbf{88.7} \\
    \bottomrule
    \end{tabular}
    }
  \label{tab:main_result}
\end{table*}

%% file: tables/compare_crf.tex
\begin{table*}
  \centering
  \caption{The $F1$ and $J$ of methods on \polnear. \textit{B.L.} and \textit{B.L.C.} denote BiLSTM and BiLSTM+CRF respectively.}
  \small{
    \begin{tabular}{l|rrr|rrr|rrr}
    \toprule
    \multicolumn{1}{c|}{\multirow{2}[4]{*}{Model}} & \multicolumn{3}{c|}{Source} & \multicolumn{3}{c|}{Cue} & \multicolumn{3}{c}{Content} \\
\cmidrule{2-10}    \multicolumn{1}{l|}{} & \multicolumn{1}{c}{$F1$-E.} & \multicolumn{1}{c}{$F1$-B.} & \multicolumn{1}{c|}{$J$} & \multicolumn{1}{c}{$F1$-E.} & \multicolumn{1}{c}{$F1$-B.} & \multicolumn{1}{c|}{$J$} & \multicolumn{1}{c}{$F1$-E.} & \multicolumn{1}{c}{$F1$-B.} & \multicolumn{1}{c}{$J$} \\
    \midrule
    CNN              & 52.7  & 65.9  & 45.1  & 58.4  & 67.8  & 49.4  & 16.2  & 60.6  & 30.2 \\
    \ w. CRF & +8.3   & \textbf{+4.1} & +8.0   & \textbf{+4.3} & \textbf{+2.2} & \textbf{+3.6} & +25.8  & +1.9   & +19.3 \\
    \ w. Cofe & \textbf{+9.4} & +3.9   & \textbf{+8.1} & +3.7   & +2.1   & +3.2   & \textbf{+31.8} & \textbf{+3.1} & \textbf{+21.9} \\
    \midrule
    GRU              & 46.5  & 58.2  & 36.7  & 59.1  & 68.1  & 48.8  & 51.3  & 65.0  & 51.3 \\
    \ w. CRF & +19.3  & +13.7  & +19.3  & +6.2   & +3.9   & +6.8   & +3.8   & +0.8   & \textbf{+6.2} \\
    \ w. Cofe & \textbf{+20.5} & \textbf{+14.6} & \textbf{+19.7} & \textbf{+7.2} & \textbf{+4.6} & \textbf{+7.5} & \textbf{+6.9} & \textbf{+1.9} & +6.2 \\
    \midrule
    LSTM             & 46.1  & 56.4  & 35.7  & 58.6  & 67.5  & 47.9  & 50.4  & 65.5  & 50.8 \\
    \ w. CRF & +19.4  & +14.7  & +19.4  & +6.4   & +4.2   & +6.7   & +4.6   & +0.3   & +5.4 \\
    \ w. Cofe & \textbf{+21.8} & \textbf{+16.3} & \textbf{+20.9} & \textbf{+6.5} & \textbf{+4.3} & \textbf{+7.1} & \textbf{+7.6} & \textbf{+0.7} & \textbf{+6.0} \\
    \midrule
    BiLSTM & 64.1  & 74.4  & 56.8  & 63.3  & 72.6  & 55.1  & 53.4  & 67.3  & 53.7 \\
    \ w. CRF & +5.5   & +1.3   & +4.5   & +3.4   & +1.2   & +2.6   & +5.6   & +2.1   & +6.6 \\
    \ w. Cofe & \textbf{+7.1} & \textbf{+3.7} & \textbf{+7.0} & \textbf{+3.7} & \textbf{+1.3} & \textbf{+3.4} & \textbf{+8.8} & \textbf{+3.4} & \textbf{+9.1} \\
    \midrule
    BERT  & 81.1  & 86.2  & 74.8  & 74.0  & 81.1  & 67.4  & 68.9  & 78.7  & 70.0 \\
    \ w. CRF & +1.1   & +0.3   & +0.8   & +0.9   & +0.9   & +1.5   & +2.1   & +0.2   & +2.8 \\
    \ w. CNN & -0.3  & +0.6   & +0.5   & +0.0   & +1.0   & +1.2   & +0.7   & +0.3   & +0.8 \\
    \ w. LSTM & +0.5   & +0.4   & +0.4   & -0.3  & 0.0   & +0.1   & +2.0   & +0.3   & +1.0 \\
    \ w. \textit{B.L.} & -0.6  & -0.1  & -0.5  & -0.5  & +0.7   & +0.5   & +0.7   & -0.2  & -0.6 \\
    \ w. \textit{B.L.C.} & +1.4  & +0.3  & +1.2  & \textbf{+1.4}  & +0.9  & +1.8   & +2.9 & +0.2  & +2.4 \\
    \ w. Cofe & \textbf{+2.2} & \textbf{+0.9} & \textbf{+1.7} & +1.3 & \textbf{+1.2} & \textbf{+2.0} & \textbf{+4.0} & \textbf{+1.0} & \textbf{+3.2} \\
    \bottomrule
    \end{tabular}
  }
  \label{tab:compare_crf}
\end{table*}

%% file: tables/ablation.tex
\begin{table*}
  \centering
  \caption{Ablation study on \polnear dataset.}
    \small{
    \begin{tabular}{l|rrr|rrr|rrr}
    \toprule
    \multicolumn{1}{c|}{\multirow{2}[4]{*}{Model}} & \multicolumn{3}{c|}{Source} & \multicolumn{3}{c|}{Cue} & \multicolumn{3}{c}{Content} \\
\cmidrule{2-10}    \multicolumn{1}{l|}{} & \multicolumn{1}{c}{$F1$-E.} & \multicolumn{1}{c}{$F1$-B.} & \multicolumn{1}{c|}{$J$} & \multicolumn{1}{c}{$F1$-E.} & \multicolumn{1}{c}{$F1$-B.} & \multicolumn{1}{c|}{$J$} & \multicolumn{1}{c}{$F1$-E.} & \multicolumn{1}{c}{$F1$-B.} & \multicolumn{1}{c}{$J$} \\
    \midrule
    CofeNet & 83.2  & 87.1  & 76.4  & 75.3  & 82.3  & 69.4  & 72.9  & 79.6  & 73.2 \\
    \quad w.o. g.m. & -1.0  & -0.6  & -0.9  & -0.2  & -0.2  & -1.0  & -0.8  & -0.3  & -1.2 \\
    \quad w.o. a.m. & -0.9  & -1.4  & -1.5  & -0.2  & -1.0  & -1.3  & -1.2  & -0.8  & -1.3 \\
    \quad w.o. f.l. & -2.4  & -0.8  & -1.5  & -1.9  & -0.5  & -1.5  & -2.5  & -0.3  & -2.7 \\
    \quad w.o. f.w. & -0.9  & -0.6  & -1.1  & -0.1  & -0.3  & -0.9  & -1.3  & -0.8  & -1.1 \\
    \quad w.o. c.w. & -2.0  & -1.4  & -2.0  & -1.1  & -1.0  & -1.6  & -1.4  & -1.2  & -1.2 \\
    \quad w.o. l.w. & -1.0  & -0.9  & -1.2  & -0.4  & -0.4  & -0.6  & -1.7  & -1.4  & -1.0 \\
    \bottomrule
    \end{tabular}
    }
  \label{tab:ablation}
\end{table*}

%% file: tables/model-setting.tex
\begin{table}[h]
  \centering
  \caption{
    CofeNet-BERT experimental configuration on \polnear, \riqua and \datasetzh datasets. The sampling ratio is the value selection ratio of the former label during training. The three values represent the proportions of truth label, predict label and random label.
  }
  \small{
    \begin{tabular}{c|ccc}
    \toprule
    \multicolumn{2}{c}{\textit{Training hyperparameters}} \\
    \midrule
    Optimizer  & Adam \\
    Learning rate except BERT  & 1e-3  \\
    Learning rate of BERT  & 5e-5   \\
    
    \midrule
    \multicolumn{2}{c}{\textit{The hyperparameters of BERT}} \\
    \midrule
    Encoder layer  & 12  \\
    Attention head  & 12 \\
    Hidden size  & 768   & \\
    Intermediate size  & 3,072   \\
    \midrule
    \multicolumn{2}{c}{\textit{The hyperparameters of \modelname}} \\
    \midrule
    Hidden size  & 100  \\
    Label embedding & 100\\
    Number of Former labels $k$& 1 \\
    Number of Former words $n$ & 3 \\
    Number of Latter words $m$ & 3 \\
    \bottomrule
    \end{tabular}%
    }
  \label{tab:hyper-param}%
\end{table}